\documentclass[11pt,a4paper]{article}
\usepackage{acl2015}
\usepackage{url}
\usepackage{latexsym}

\usepackage{booktabs} % For formal tables

\usepackage{times}
\usepackage{latexsym}
\usepackage{hyperref}
\usepackage{times}
\usepackage{graphicx}
\usepackage[utf8]{inputenc}
\usepackage{amsmath}
\usepackage[T1]{fontenc}
\usepackage{tipa}
\usepackage{wrapfig}
\usepackage{enumitem}
\usepackage[ruled]{algorithm2e}

\title{Unsupervised Separation of Transliterable and\\  Native Words for Malayalam}

\author{Deepak P \\
Queen's University Belfast, UK \\
{\tt deepaksp@acm.org} \\}

\date{}

\begin{document}
\maketitle
\begin{abstract}
Differentiating intrinsic language words from transliterable words is a key step aiding text processing tasks involving different natural languages. We consider the problem of unsupervised separation of transliterable words from native words for text in Malayalam language. Outlining a key observation on the diversity of characters beyond the word stem, we develop an optimization method to score words based on their nativeness. Our method relies on the usage of probability distributions over character n-grams that are refined in step with the nativeness scorings in an iterative optimization formulation. Using an empirical evaluation, we illustrate that our method, DTIM, provides significant improvements in nativeness scoring for Malayalam, establishing DTIM as the preferred method for the task. 
\end{abstract}

\maketitle

\section{Introduction}

Malayalam is an agglutinative language from the southern Indian state of Kerala where it is the official state language. It is spoken by 38 million native speakers, three times as many speakers as Hungarian~\cite{vincze2013dependency} or Greek~\cite{ntoulas2001using}, for which specialized techniques have been developed in other contexts. The growing web presence of Malayalam necessitates automatic techniques to process Malayalam text. A major hurdle in harnessing Malayalam text from social and web media for multi-lingual retrieval and machine translation is the presence of a large amount of transliterable words. By transliterable words, we mean both (a) words (from English) like {\it police} and {\it train} that virtually always appear in transliterated form in contemporary Malayalam, and (b) proper nouns such as names that need to be transliterated than translated to correlate with English text. On a manual analysis of a news article dataset, we found that transliterated words and proper nouns each form $10$-$12\%$ of all distinct words. It is useful to transliterate such words for scenarios that involve processing Malayalam text in the company of English text;  this will avoid them being treated as separate index terms (wrt their transliteration) in a multi-lingual retrieval engine, and help a statistical translation system to make use of the link to improve effectiveness. In this context, it ia notable that there has been recent interest in devising specialized methods to translate words that fall outside the core vocabulary~\cite{DBLP:conf/acl/TsvetkovD15}. %We arrived at the problem in the processing of Malayalam medical diagnostics, a realm with abundant usage of english-origin medical terminology, where the task involves a human-in-the-loop semi-automatic transliteration. 
%learn word/phrase associations more accurately. %

In this paper, we consider the problem of separating out such {\it transliterable} words from the other words within an unlabeled dataset; we refer to the latter as {\it ``native''} words. We propose an {\it unsupervised} method, DTIM, that takes a dictionary of distinct words from a Malayalam corpus and scores each word based on their {\it nativeness}. Our optimization method, DTIM, iteratively refines the nativeness scoring of each word, leveraging dictionary-level statistics modelled using character n-gram probability distributions. Our empirical analysis establishes the effectiveness of DTIM. %the proposed method.
%that our method is able to achieve significantly higher accuracy vis-a-vis methods from literature.%, for the task of separating transliterable words from pure ones. We draw upon a key observation about the nature of native word stems, and develop an EM-based optimization method. 

We outline related work in the area in Section~\ref{sec:related}. This is followed by the problem statement in Section~\ref{sec:probdef} and the description of our proposed approach in Section~\ref{sec:method}. Our empirical analysis forms Section~\ref{sec:expts} followed by conclusions in Section~\ref{sec:conclusions}. 

\section{Related Work}\label{sec:related}

Identification of transliterable text fragments, being a critical task for cross-lingual text analysis, has attracted attention since the 1990s. While most methods addressing the problem have used supervised learning, there have been some methods that can work without labeled data. We briefly survey both classes of methods. 
%We briefly summarize methods for the task of identifying transliterable text fragments. 
%We now briefly summarize related literature. 

%{\bf Identifying Transliterable Words/Phrases:} 

\subsection{ Supervised and `pseudo-supervised' Methods}

%\noindent{\bf Supervised and `pseudo-supervised' Methods:} 

An early work\cite{chen1996identification} focuses on a sub-problem, that of supervised identification of proper nouns for Chinese. \cite{jeong1999automatic} consider leveraging decision trees to address the related problem of learning transliteration and back-transliteration rules for English/Korean word pairs. Recognizing the costs of procuring training data, \cite{baker2008statistical} and \cite{goldberg2008identification} explore usage of pseudo-transliterable words generated using transliteration rules on an English dictionary for Korean and Hebrew respectively. Such pseudo-supervision, however, would not be able to generate uncommon domain-specific terms such as medical/scientific terminology for usage in such domains (unless specifically tuned), and is hence limited in utility. %We draw inspiration from the usage of character-level language models in their method, and we will harness such models in our unsupervised method too. 

\subsection{Unsupervised Methods}\label{sec:unsupervised}

A recent work proposes that multi-word phrases in Malayalam text where their component words exhibit strong co-occurrence be categorized as transliterable phrases \cite{prasad2014technique}. Their intuition stems from observing contiguous words such as {\it test dose} which often occur in transliterated form while occurring together, but get replaced by native words in other contexts. Their method is however {\it unable to identify single transliterable words}, or phrases involving words such as {\it train} and {\it police} whose transliterations are heavily used in the company of native Malayalam words. A recent method for Korean~\cite{koo2015unsupervised} starts by identifying a seed set of transliterable words as those that begin or end with consonant clusters and have vowel insertions; this is specific to Korean since Korean words apparently do not begin or end with consonant clusters. High-frequency words are then used as seed words for native Korean for usage in a Naive Bayes classifier. In addition to the outlined reasons that make both the unsupervised methods inapplicable for our task, they both presume availability of corpus frequency statistics. We focus on a general scenario assuming the availability of only a word lexicon.%, motivated by a use case in Kerala's Health Services where corpus statistics are indicative of disease distribution and considered confidential. 
%The scenario is particularly suited for our work due to abundant usage of non-native words in the medical domain. 

%corpus statistics are hard to get due to concerns about localized disease distribution getting revealed in the process.

%, without frequency or co-occurrence statistics. 

\subsection{Positioning the Transliterable Word Identification Task} 

Nativeness scoring of words may be seen as a vocabulary stratification step (upon usage of thresholds) for usage by downstream applications. A multi-lingual text mining application that uses Malayalam and English text would benefit by transliterating non-native Malayalam words to English, so the transliterable Malayalam token and its transliteration is treated as the same token. For machine translation, transliterable words may be channeled to specialized translation methods (e.g.,~\cite{DBLP:conf/acl/TsvetkovD15}) or for manual screening and translation. 

\section{Problem Definition}\label{sec:probdef}

We now define the problem more formally. Consider $n$ distinct words obtained from Malayalam text, $\mathcal{W} = \{ \ldots, w, \ldots \}$. Our task is to devise a technique that can use $\mathcal{W}$ to arrive at a {\it nativeness} score for each word, $w$, within it, as $w_n$. We would like $w_n$ to be an accurate quantification of native-ness of word $w$. For example, when words in $\mathcal{W}$ are ordered in the decreasing order of $w_n$ scores, we expect to get the native words at the beginning of the ordering and vice versa. We do not presume availability of any data other than $\mathcal{W}$; this makes our method applicable across scenarios where corpus statistics are unavailable due to privacy or other reasons. 

\subsection{Evaluation}

Given that it is easier for humans to crisply classify each word as either native or transliterable (nouns or transliterated english words) in lieu of attaching a score to each word, the nativeness scoring (as generated by a scoring method such as ours) often needs to be evaluated against a crisp nativeness assessment, i.e., a scoring with scores in $\{0, 1\}$. To aid this, we consider the ordering of words in the labeled set in the decreasing (or more precisely, non-increasing) order of {\it nativeness} scores (each method produces an ordering for the dataset). To evaluate this ordering, we use two sets of metrics for evaluation: 

%\vspace{-0.1in}
\begin{itemize}%[leftmargin=*]
\setlength\itemsep{0in}
\item {\it Precision at the ends of the ordering:} {\bf Top-k precision} denotes the fraction of {\it native} words within the $k$ words at the {\it top} of the ordering; analogously, {\bf Bottom-k precision} is the fraction of {\it transliterable} words among the {\it bottom k}. Since a good scoring would likely put native words at the top of the ordering and the transliterable ones at the bottom, a good scoring method would intuitively score high on both these metrics. We call the average of the top-k and bottom-k precision for a given k, as {\bf Avg-k precision}. These measures, evaluated at varying values of $k$, indicate the quality of the nativeness scoring. 
% words.% at the {\it bottom}. 
\item {\it Clustering Quality:} Consider the cardinalities of the native and transliterable words from the labeled set as being $N$ and $T$ respectively. We now take the top-N words and bottom-T words from the ordering generated by each method, and compare against the respective labeled sets as in the case of standard clustering quality evaluation\footnote{{\scriptsize \url{https://nlp.stanford.edu/IR-book/html/htmledition/evaluation-of-clustering-1.html}}}. Since the cardinalities of the generated native (transliterable) cluster and the native (transliterable) labeled set is both $N$ ($T$), the {\bf Recall} of the cluster is identical to its {\bf Purity/Precision}, and thus, the {\bf F-measure} too; we simply call it {\bf Clustering Quality}. A cardinality-weighted average of the clustering quality across the native and transliterable clusters yields a single value for the clustering quality across the dataset. It may be noted that we are not making the labeled dataset available to the method generating the ordering, instead merely using it's cardinalities for evaluation purposes. 
\end{itemize}

%and the transliterable words at the end. 

%

%We will evaluate the quality of the scoring using appropriate precision metrics at the top-$k$ and bottom-$k$, for varying values of $k$. 

\section{Our Method: DTIM}\label{sec:method}

We now introduce our method, {\bf D}iversity-based {\bf T}ransliterable Word {\bf I}dentification for {\bf M}alayalam (DTIM). We use probability distributions over character n-grams to separately model transliterable and native words, and develop an optimization framework that alternatively refines the n-gram distributions and nativeness scoring within each iteration. DTIM involves an initialization that induces a ``coarse'' separation between native and transliterable words followed by iterative refinement. The initialization is critical in optimization methods that are vulnerable to local optima; the pure word distribution needs to be initialized to ``coarsely'' prefer pure words over transliterable words. This will enable further iterations to exploit the initial preference direction to further refine the model to ``attract'' the pure words more strongly and weaken any initial preference to transliterable words. The vice versa holds for the transliterable word models. We will first outline the initialization step followed by the description of the method. 

%; we develop an EM method that refine them iteratively so they characterize their respective sets better. DTIM involves an initialization that induces a ``coarse'' separation between native and transliterable words followed by iterative refinement. %We will first outline the initialization step followed by the description of the method. 

%

\subsection{Diversity-based Initialization}\label{sec:divinit}

%, and the difference in the quantum of such variety across transliterable and pure words. 

Our initialization is inspired by an observation on the variety of suffixes attached to a word stem. Consider a word stem {\it |pu|ra|}\footnote{{\scriptsize We will represent Malayalam words in transliterated form for reading by those who might not be able to read Malayalam. A pipe would separate Malayalam characters; for example |pu| corresponds to a single Malayalam character. }}, a stem commonly leading to native Malayalam words; its suffixes are observed to start with a variety of characters such as {\it |ttha|} (e.g., {\it |pu|ra|ttha|kki|}), {\it |me|} (e.g., {\it |pu|ra|me|}), {\it |mbo|} (e.g., {\it |pu|ra|mbo|kku|}) and {\it |ppa|} (e.g., {\it |pu|ra|ppa|du|}). On the other hand, stems that mostly lead to transliterable words often do not exhibit so much of diversity. For example, {\it |re|so|} is followed only by {\it |rt|} (i.e., {\it resort}) and {\it |po|li|} is usually only followed by {\it |s|} (i.e., {\it police}). Some stems such as {\it |o|ppa|} lead to transliterations of two English words such as {\it open} and {\it operation}. Our observation, upon which we model the initialization, is that the variety of suffixes is generally correlated with native-ness (i.e., propensity to lead to a native word) of word stems. This is intuitive since non-native word stems provide limited flexibility to being modified by derivational or inflectional suffixes as compared to native ones. 

For simplicity, we use the first two characters of each word as the word stem; we will evaluate the robustness of DTIM to varying stem lengths in our empirical evaluation, while consistently using the stem length of two characters in our description. We start by associating each distinct word stem in $\mathcal{W}$ with the number of unique third characters that follow it (among words in $\mathcal{W}$); in our examples, {\it |pu|ra|} and {\it |o|pa|} would be associated with $4$ and $2$ respectively. We initialize the {\it native-ness} weights as proportional to the diversity of 3$^{rd}$ characters beyond the stem:

%\begin{wrapfigure}{l}{.47\textwidth}
%\vspace{-0.2in}
\begin{equation}\label{eq:init}
w_{n_{0}} = min\bigg\{ 0.99, \frac{|u3(w_{stem},\mathcal{W})|}{\tau}\bigg\}
\end{equation}
%%%\vspace{-0.5in}
%\end{wrapfigure}

\noindent where $u3(w_{stem},\mathcal{W})$ denotes the set of third characters that follow the stem of word $w$ among words in $\mathcal{W}$. We flatten off $w_{n_0}$ scores beyond a diversity of $\tau$ (note that a diversity of $\tau$ or higher will lead to the second term becoming $1.0$ or higher, kicking in the min function to choose $0.99$ for $w_{n_0}$) as shown in the above equation. We give a small transliterable-ness weight even to highly diverse stems to reduce over-reliance on the initialization. We set $\tau = 10$ based on our observation from the dataset that most word stems having more than $10$ distinct characters were seen to be native. As in the case of word stem length, we study DTIM trends across varying $\tau$ in our empirical analysis. $w_{n_0}$ is in $[0,1]$; analogously, $(1-w_{n_0})$ may be regarded as a score of transliterable-ness. %This initialization based on diversity after the stem lends the name to the technique. %The word-stem length may be varied with corresponding variations to the initialization heuristic; with a word-stem length of $3$ characters, Equation~\ref{eq:init} would use $u4(.,.)$ instead of $u3(.,.)$. We will analyze performance across variations in word stem length and $\tau$ in our empirical analysis. 

\subsection{Objective Function and Optimization Framework}

As outlined earlier, we use separate character n-gram probability distributions to model native and transliterable words. We would like these probability distributions support the nativeness scoring, and vice versa. While the size of the n-grams (i.e., whether $n=1,2,3$ or $4$) is a system-level parameter, we use $n=1$ for simplicity in our description. We denote the native and transliterable distributions as $\mathcal{N}$ and $\mathcal{T}$ respectively, with $\mathcal{N}(c)$ and $\mathcal{T}(c)$ denoting the weight associated with the character $c$ according to the distributions. Consider the following function, given a particular state for the $\mathcal{N}$, $\mathcal{T}$ and $w_n$s:

\begin{equation}\label{eq:maximizing}
\prod_{w \in \mathcal{W}} \prod_{c \in w} \bigg(  w_n^2 \times \mathcal{N}(c) + (1-w_n)^2 \times \mathcal{T}(c) \bigg)
\end{equation}

This measures the aggregate supports for words in $\mathcal{W}$, the support for each word measured as an interpolated support from across the distributions $\mathcal{N}$ and $\mathcal{T}$ with weighting factors squares of the nativeness scores (i.e., $w_n$s) and transliterableness scores (i.e., $(1-w_n)$s) respectively. Similar mixing models have been used earlier in emotion lexicon learning~\cite{bandhakavi2014generating} and solution post discovery~\cite{deepak2014unsupervised}. The squares of the nativeness scores are used in our model (instead of the raw scores) for optimization convenience. A highly native word should intuively have a high $w_n$ (nativeness) and a high support from $\mathcal{N}$ and correspondingly low transliterable-ness (i.e., $(1-w_n)$) and support from $\mathcal{T}$; a highly transliterable word would be expected to have exactly the opposite. Due to the design of Eq.~\ref{eq:maximizing} in having the higher terms multiplied with each other (and so for the lower terms), this function would be maximized for a desirable estimate of the variables $\theta = \{ \mathcal{N}, \mathcal{T}, \{ \ldots, w_n, \ldots \} \}$. Conversely, by striving to maximize the objective function, we would arrive at a desirable estimate of the variables. An alternative construction yielding a minimizing objective would be as follows:

\begin{equation}\label{eq:minimizing}
\prod_{w \in \mathcal{W}} \prod_{c \in w} \bigg(  (1-w_n)^2 \times \mathcal{N}(c) + w_n^2 \times \mathcal{T}(c) \bigg)
\end{equation}

In this form, given a good estimate of the variables, the native (transliterable) words have their nativeness (transliterableness) weights multiplied with the support from the transliterable (native) models. In other words, maximizing the objective in Eq. \ref{eq:maximizing} is semantically similar to minimizing the objective in Eq.~\ref{eq:minimizing}. As we will illustrate soon, it is easier to optimize for $\mathcal{N}$ and $\mathcal{T}$ using the maximizing formulation in Eq.~\ref{eq:maximizing} while the minimizing objective in Eq.~\ref{eq:minimizing} yields better to optimize for the word nativeness scores, $\{ \ldots, w_n, \ldots \}$. 

\subsection{Learning $\mathcal{N}$ and $\mathcal{T}$ using the Maximizing Objective}

We start by taking the log-form of the objective in Eq.~\ref{eq:maximizing} (this does not affect the optimization direction), yielding:

\begin{equation}\label{eq:maximizinglog}
\mathcal{O}_{max} = \sum_{w \in \mathcal{W}} \sum_{c \in w} ln\ \bigg(  w_n^2 \times \mathcal{N}(c) + (1-w_n)^2 \times \mathcal{T}(c) \bigg)
\end{equation}

The distributions, being probability distributions over n-grams, should sum to zero. This constraint, for our unigram models, can be written as:

\begin{equation}
\sum_{c} \mathcal{N}(c) = \sum_{c} \mathcal{T}(c) = 1
\end{equation}

Fixing the values of $\{ \ldots, w_n, \ldots \}$ and $\mathcal{T}$ (or $\mathcal{N}$), we can now identify a better estimate for $\mathcal{N}$ (or $\mathcal{T}$) by looking for an optima (i.e., where the objective function has a slope of zero). Towards that, we take the partial derivative (or slope) of the objective for a particular character. :

\begin{equation}
\frac{\partial \mathcal{O}_{max}}{\partial \mathcal{N}(c')} = \bigg( \sum_{w \in \mathcal{W}} \frac{freq(c',w) \times w_n^2}{\big( w_n^2 \mathcal{N}(c') + (1-w_n)^2 \mathcal{T}(c')\big)} \bigg) + \lambda_{\mathcal{N}} 
\end{equation}

where $freq(c',w)$ is the frequency of the character $c'$ in $w$ and $\lambda_{\mathcal{N}}$ denotes the Lagrangian multiplier corresponding to the sum-to-unity constraint for $\mathcal{N}$. Equating this to zero does not however yield a closed form solution for $\mathcal{N}'$, but a simple re-arrangement yields an iterative update formula:

\begin{equation}\label{eq:nupdate}
\mathcal{N}(c') \propto \sum_{w \in \mathcal{W}} \frac{freq(c',w) \times w_n^2}{\big( w_n^2 + (1-w_n)^2 \frac{\mathcal{T}(c')}{\mathcal{N}_P(c')}\big)}
\end{equation}

The $\mathcal{N}$ term in the RHS is denoted as $\mathcal{N}_P$ to indicate the usage of the previous estimate of $\mathcal{N}$. The sum-to-one constraint is trivially achieved by first estimating the $\mathcal{N}(c')$s by treating Eq.~\ref{eq:nupdate} as equality, followed by normalizing the scores across the character vocabulary. Eq.~\ref{eq:nupdate} is intuitively reasonable, due to establishing a somewhat direct relationship between $\mathcal{N}$ and $w_n$ (in the numerator), thus allowing highly native words to contribute more to building $\mathcal{N}$. The analogous update formula for $\mathcal{T}$ fixing $\mathcal{N}$ turns out to be:

\begin{equation}\label{eq:tupdate}
\mathcal{T}(c') \propto \sum_{w \in \mathcal{W}} \frac{freq(c',w) \times (1-w_n)^2}{\big( (1-w_n)^2 + w_n^2 \frac{\mathcal{N}(c')}{\mathcal{T}_P(c')}\big)}
\end{equation}

Eq.~\ref{eq:nupdate} and Eq.~\ref{eq:tupdate} would lead us closer to a maxima for Eq.~\ref{eq:maximizinglog} is their second (partial) derivatives are negative\footnote{\url{http://mathworld.wolfram.com/SecondDerivativeTest.html}}. To verify this, we note that the second (partial) derivative wrt $\mathcal{N}(c')$ is as follows

\begin{equation*}
\frac{\partial^2 \mathcal{O}_{max}}{\partial^2 \mathcal{N}(c')} = \hspace{3in} 
\end{equation*}

\vspace{-0.3in}

\begin{equation}
(-1) \times \sum_{w \in \mathcal{W}} \frac{freq(c',w) (w_n^2)^2}{\big( w_n^2 \mathcal{N}(c') + (1-w_n)^2 \mathcal{T}(c')\big)^2}
\end{equation}

It is easy to observe that the RHS is a product of $-1$ and a sum of a plurality of positive terms (square terms that are trivially positive, with the exception being the $freq(.,.)$ term which is also non-negative by definition), altogether yielding a negative value. That the the second (partial) derivative is negative confirms that the update formula derived from the first partial derivative indeed helps in maximizing $\mathcal{O}_{max}$ wrt $\mathcal{N}(c')$. A similar argument holds for the $\mathcal{T}(c')$ updates as well, which we omit for brevity. 

\subsection{Learning the nativeness scores, $\{ \ldots, w_n, \ldots \}$, using the Minimizing Objective}

Analogous to the previous section, we take the log-form of Eq.~\ref{eq:minimizing}:

\begin{equation}\label{eq:minimizinglog}
\mathcal{O}_{min} = \sum_{w \in \mathcal{W}} \sum_{c \in w} ln\ \bigg(  (1-w_n)^2 \times \mathcal{N}(c) + w_n^2 \times \mathcal{T}(c) \bigg)
\end{equation}

Unlike the earlier case, we do not have any constraints since the sum-to-unit constraint on the nativeness and transliterableness scores are built in into the construction. We now fix the values of all other variables and find the slope wrt $w'_n$, where $w'$ indicates a particular word in $\mathcal{W}$. 

\begin{equation}
\frac{\partial \mathcal{O}_{min}}{\partial w'_n} = \sum_{c \in w'} \frac{2 w'_n \mathcal{T}(c) + 2 w'_n \mathcal{N}(c)  - 2 \mathcal{N}(c)}{\big(w'^2_n \mathcal{T}(c)  + (1-w'_n)^2 \mathcal{N}(c)\big)}
\end{equation}

We equate the slope to zero and form an iterative update formula, much like in the distribution estimation phase. 

\begin{equation}\label{eq:weightupdate}
w'_n = \frac{\sum_{c \in w'} \frac{\mathcal{N}(c)}{(w'^2_n \mathcal{T}(c)  + (1-w'_n)^2 \mathcal{N}(c) )}}{\sum_{c \in w'} \frac{\mathcal{N}(c) + \mathcal{T}(c)}{(w'^2_n \mathcal{T}(c)  + (1-w'_n)^2 \mathcal{N}(c) )}}
\end{equation}

Using the previous estimates of $w'_n$ for the RHS yields an iterative update form for the nativeness scores. As in the model estimation phase, the update rule establishes a reasonably direct relationship between $w'_n$ and $\mathcal{N}$. Since our objective is to minimize $\mathcal{O}_{min}$, we would like to verify the direction of optimization using the second partial derivative. 

\begin{equation*}
\frac{\partial^2 \mathcal{O}_{min}}{\partial^2 w'_n} = \hspace{3in}
\end{equation*}

\vspace{-0.3in}

\begin{equation}
\sum_{c \in w'} \frac{\mathcal{N}(c) \mathcal{T}(c) - \big(w'_n \mathcal{T}(c) - (1-w'_n) \mathcal{N}(c)\big)^2}{\big(w'^2_n \mathcal{T}(c)  + (1-w'_n)^2 \mathcal{N}(c)\big)^2}
\end{equation}

We provide an informal argument for the positivity of the second derivative; note that the denominator is a square term making it enough to analyze just the numerator term. Consider a highly native word (high $w'_n$) whose characters would intuitively satisfy $\mathcal{N}(c) > \mathcal{T}(c)$. For the boundary case of $w'_n = 1$, the numerator term reduces to $\mathcal{T}(c) \times (\mathcal{N}(c) - \mathcal{T}(c))$ which would be positive given the expected relation between $\mathcal{N}(c)$ and $\mathcal{T}(c)$. A similar argument holds for highly transliterable words. For words with $w'_n \rightarrow 0.5$ where we would expect $\mathcal{N}(c) \approx \mathcal{T}(c)$, the numerator becomes $\mathcal{N}(c) \mathcal{T}(c) - 0.25 (\mathcal{T}(c) - \mathcal{N}(c))^2$, which is expected to be positive since the difference term is small, making it's square very small in comparison to the first product term. To outline the informal nature of the argument, it may be noted that $\mathcal{T}(c) > \mathcal{N}(c)$ may hold for certain characters within highly native words; but as long as most of the characters within highly native words satisfy the $\mathcal{N}(c) > \mathcal{T}(c)$, there would be sufficient positivity to offset the negative terms induced with such outlier characters. 

\begin{algorithm}
\DontPrintSemicolon % Some LaTeX compilers require you to use \dontprintsemicolon instead
\KwIn{A set of Malayalam words, $\mathcal{W}$}
\KwOut{A nativeness scoring $w_n \in [0,1]$ for every word $w$ in $\mathcal{W}$}
{\bf Hyper-parameters:} word stem length, $\tau$, $n$\;
Initialize the $w_n$ scores for each word using the diversity metric in Eq.~\ref{eq:init} using the chosen stem length and $\tau$\;
\While{\text{not converged and number of iterations} \text{not reached}}{
  \text{Estimate n-gram distributions} $\mathcal{N}$ and $\mathcal{T}$ using Eq.~\ref{eq:nupdate} and Eq.~\ref{eq:tupdate} respectively\;
  \text{Learn nativeness weights for each word} \text{using Eq.~\ref{eq:weightupdate}} \;
}
\Return{latest estimates of nativeness weights}\;
\caption{{\sc DTIM}}
\label{algo:dtim}
\end{algorithm}

\subsection{DTIM: The Method}

Having outlined the learning steps, the method is simply an iterative usage of the learning steps as outlined in Algorithm~\ref{algo:dtim}. In the first invocation of the distribution learning step where previous estimates are not available, we simply assume a uniform distribution across the n-gram vocabulary for usage as the previous estimates. Each of the update steps are linear in the size of the dictionary, making DTIM a computationally light-weight method. Choosing $n=2$ instead of unigrams (as used in our narrative) is easy since that simply involves replacing the $c \in w$ all across the update steps by $[c_i, c_{i+1}] \in w$ with $[c_i, c_{i+1}]$ denoting pairs of contiguous characters within the word; similarly, $n=3$ involves usage of contiguous character triplets and correspondingly learning the distributions $\mathcal{N}$ and $\mathcal{T}$ over triplets. The DTIM structure is evidently inspired by the Expectation-Maximization framework~\cite{10.2307/2984875} involving alternating optimizations of an objective function; DTIM, however, uses different objective functions for the two steps for optimization convenience.

\section{Experiments}\label{sec:expts}

We now describe our empirical study of DTIM, starting with the dataset and experimental setup leading on to the results and analyses. 

\subsection{Dataset}

We evaluate DTIM on a set of $65068$ distinct words from across news articles sourced from {\it Mathrubhumi}\footnote{{\scriptsize \url{http://www.mathrubhumi.com}}}, a popular Malayalam newspaper; this word list is made available publicly\footnote{{\scriptsize Dataset: \url{https://goo.gl/DOsFES}}}. For evaluation purposes, we got a random subset of $1035$ words labeled by one of three human annotators; that has been made available publicly\footnote{{\scriptsize Labeled Set: \url{https://goo.gl/XEVLWv}}} too, each word labeled as either {\it native}, {\it transliterable} or {\it unknown}. There were approximately 3 native words for every transliterable word in the labeled set, reflective of distribution in contemporary Malayalam usage as alluded to in the introduction. We will use the whole set of $65068$ words as input to the method, while the evaluation would obviously be limited to the labelled subset of $1035$ words. 

%There were $\approx$75\% native and $\approx25\%$ transliterable words in the labeled set, reflective of the the distribution in contemporary Malayalam.% of Malayalam. 

\subsection{Baselines} 

As outlined in Section~\ref{sec:related}, the unsupervised version of the problem of telling apart native and transliterable words for Malayalam and/or similar languages has not been addressed in literature, to the best of our knowledge. The unsupervised Malayalam-focused method\cite{prasad2014technique} (Ref: Sec~\ref{sec:unsupervised}) is able to identify only transliterable word-pairs, making it inapplicable for contexts such as a health data scenario where individual english words are often transliterated for want of a suitable malayalam alternative. The Korean method\cite{koo2015unsupervised} is too specific to Korean language and cannot be used for other languages due to the absence of a generic high-precision rule to identify a seed set of transliterable words. With both the unsupervised state-of-the-art approaches being inapplicable for our task, we compare against an intuitive generalization-based baseline, called {\bf GEN}, that orders words based on their support from the combination of a unigram and bi-gram character language model learnt over $\mathcal{W}$; this leads to a scoring as follows:

\begin{equation*}
w_n^{GEN} = \hspace{3in}
\end{equation*}

\vspace{-0.3in}

\begin{equation}\label{eq:gen}
\prod_{[c_i, c_{i+1}] \in w} \lambda \times B_{\mathcal{W}}(c_{i+1}|c_i) + (1-\lambda) \times U_{\mathcal{W}}(c_{i+1})
\end{equation}

where $B_{\mathcal{W}}$ and $U_{\mathcal{W}}$ are bigram and unigram character-level language models built over all words in $\mathcal{W}$. We set $\lambda = 0.8$~\cite{IR-548}. We experimented with higher-order models in {\bf GEN}, but observed drops in evaluation measures leading to us sticking to the usage of the unigram and bi-gram models. The form of Eq.~\ref{eq:gen} is inspired by an assumption similar to that used in both~\cite{prasad2014technique} and~\cite{koo2015unsupervised} that transliterable words are rare. Thus, we expect they would not be adequately supported by models that generalize over whole of $\mathcal{W}$. We also compare against our diversity-based initialization score from Section~\ref{sec:divinit}, which we will call as {\bf INIT}. For ease of reference, we outline the {\bf INIT} scoring:

\begin{equation}
w_n^{INIT} = min\bigg\{ 0.99, \frac{|u3(w_{stem},\mathcal{W})|}{\tau}\bigg\}
\end{equation}

The comparison against {\bf INIT} enables us to isolate and highlight the value of the iterative update formulation vis-a-vis the initialization. 

\subsection{Evaluation Measures and Setup}

As outlined in Section~\ref{sec:probdef}, we use {\it top-k}, {\it bottom-k} and {\it avg-k} precision (evaluated at varying values of $k$) as well as {\it clustering quality} in our evaluation. For the comparative evaluaton, we set DTIM parameters as the following: $\tau = 10$ and a word-stem length of $2$. We will study trends against variations across these parameters in a separate section.

\begin{table*}
\centering
\resizebox{\linewidth}{!}{%
\begin{tabular}{|c||c|c|c||c|c|c||c|c|c||c|c|c||}
		\hline
		\hline
		& \multicolumn{3} {|c|} {k=50} & \multicolumn{3} {|c|} {k=100} & \multicolumn{3} {|c|} {k=150} & \multicolumn{3} {|c|} {k=200} \\
		& Top-k & Bot-k & Avg-k & Top-k & Bot-k & Avg-k & Top-k & Bot-k & Avg-k & Top-k & Bot-k & Avg-k \\
		\hline
		\hline
		INIT & 0.88 & 0.50 & 0.69 & 0.90 & 0.40 & 0.65 & 0.90 & 0.41 & 0.66 & 0.90 & 0.38 & 0.64 \\
		GEN & 0.64 & 0.10 & 0.37 & 0.58 & 0.11 & 0.35 & 0.60 & 0.15 & 0.38 & 0.64 & 0.17 & 0.41 \\
		\hline
		DTIM (n=1) & 0.94 & 0.64 & 0.79 & 0.90 & 0.56 & 0.73 & 0.90 & 0.49 & 0.70 & 0.92 & 0.48 & 0.70 \\
		DTIM (n=2) & {\bf 1.00} & {\bf 0.78} & {\bf 0.89} & {\bf 0.94} & 0.68 & 0.81 & {\bf 0.93} & 0.57 & 0.75 & {\bf 0.95} & 0.52 & 0.74 \\
		DTIM (n=3) & 0.86 & 0.76 & 0.81 & 0.91 & {\bf 0.75} & {\bf 0.83} & 0.92 & {\bf 0.69} & {\bf 0.81} & 0.92 & 0.64 & {\bf 0.78} \\
		DTIM (n=4) & 0.82 & 0.74 & 0.78 & 0.87 & 0.69 & 0.78 & 0.83 & 0.62 & 0.73 & 0.85 & {\bf 0.65} & 0.75 \\
		\hline
		\hline
		\end{tabular}}
\caption{Top-k and Bottom-k Precision (best result in each column highlighted)}
\label{tab:prectable}
%\vspace{-0.2in}
\end{table*}

\begin{table}
\centering
\resizebox{\linewidth}{!}{%
\begin{tabular}{|c|c|c|c|}
		\hline
		\hline
		& Native &  Transliterable & Weighted Average \\
		\hline
		\hline
		INIT & 0.79 & 0.38 & 0.69 \\
		GEN & 0.73 & 0.17 & 0.59 \\
		\hline
		DTIM (n=1) & 0.81 & 0.44 & 0.72 \\
		DTIM (n=2) & 0.84 & 0.50 & 0.75 \\
		DTIM (n=3) & {\bf 0.86} & {\bf 0.60} & {\bf 0.79} \\
		DTIM (n=4) & {\bf 0.86} & {\bf 0.60} & {\bf 0.79} \\
		\hline
		\hline
		\end{tabular}}
\caption{Clustering Quality (best result in each column highlighted)}
\label{tab:recall}
%\vspace{-0.2in}
\end{table}

%\vspace{-0.2in}

\subsection{Experimental Results}

\subsubsection{Precision at the ends of the Ordering} 

Table~\ref{tab:prectable} lists the precision measures over various values of $k$. It may be noted that any instantiation of DTIM (across the four values of n-gram size, $n$) is able to beat the baselines convincingly on each metric on each value of $k$, convincingly establishing the effectiveness of the DTIM formulation. DTIM is seen to be much more effective in separating out the native and transliterable words at either ends of the ordering, than the baselines. It is also notable that EM-style iterations are able to significantly improve upon the initialization (i.e., INIT). That the bottom-k precision is seen to be consistently lower than top-k precision needs to be juxtaposed in the context of the observation that there were only $25\%$ transliterable words against $75\%$ native words; thus, the lift in precision against a random ordering is much more substantial for the transliterable words. The trends across varying n-gram sizes (i.e., $n$) in DTIM is worth noting too; the higher values of $n$ (such as $3$ and $4$) are seen to make more errors at the ends of the lists, whereas they catch-up with the $n\in \{1,2\}$ versions as $k$ increases. This indicates that smaller-n DTIM is being able to tell apart a minority of the words exceedingly well (wrt native-ness), whereas the higher n-gram modelling is able to spread out the gains across a larger spectrum of words in $\mathcal{W}$. Around $n=4$ and beyond, sparsity effects (since 4-grams and 5-grams would not occur frequently, making it harder to exploit their occurrence statistics) are seen to kick in, causing reductions in precision. 

% by up to 18 percentage points in precision. %The results also suggest that our simple diversity based initialization is much more accurate that the generalization-scoring baseline. 

%This establishes DTIM as the preferred method for the task. 

\subsubsection{Clustering Quality} 

Table~\ref{tab:recall} lists the clustering quality metric across the various methods. Clustering quality, unlike the precision metrics, is designed to evaluate the entire ordering without limiting the analysis to just the top-k and bottom-k entries. As in the earlier case, DTIM convincingly outperforms the baselines by healthy margins across all values of $n$. Consequent to the trends across $n$ as observed earlier, $n \in \{3,4\}$ are seen to deliver better accuracies, with such gains tapering off beyond $n=4$ due to sparsity effects. The words, along with the DTIM nativeness scores for $n=3$, can be viewed at \url{https://goo.gl/OmhlB3} (the list excludes words with fewer than $3$ characters).

\begin{table}
\centering
\resizebox{\linewidth}{!}{%
\begin{tabular}{|c|c|c|c|c|c|c|}
		\hline
		\hline
		$\tau \rightarrow$ & 5 & 10 & 20 & 50 & 100 & 1000 \\
		\hline
		\hline
		$n=1$ & 0.72 & 0.72 & 0.72 & 0.72 & 0.72 & 0.72 \\
		$n=2$ & 0.74 & 0.75 & 0.75 & 0.74 & 0.74 & 0.74 \\
		$n=3$ & 0.77 & 0.79 & 0.78 & 0.78 & 0.78 & 0.78 \\
		$n=4$ & 0.78 & 0.79 & 0.79 & 0.79 & 0.79 & 0.79 \\
		\hline
		\end{tabular}}
\caption{DTIM Clustering Quality against $\tau$}
\label{tab:tau}
%\vspace{-0.2in}
\end{table}

\begin{table}
\centering
\resizebox{\linewidth}{!}{%
\begin{tabular}{|c|c|c|c|c|}
		\hline
		\hline
		Stem Length $\rightarrow$ & 1 & 2 & 3 & 4 \\
		\hline
		\hline
		$n=1$ & 0.64 & 0.72 & {\bf 0.75} & 0.56 \\
		$n=2$ & 0.58 & {\bf 0.75} & 0.74 & 0.55 \\
		$n=2$ & 0.59 & {\bf 0.79} & 0.69 & 0.60 \\
		$n=2$ & 0.58 & {\bf 0.79} & 0.69 & 0.62 \\
		\hline
		\end{tabular}}
\caption{DTIM Clustering Quality against Word Stem Length (best result in each row highlighted)}
\label{tab:stemlength}
%\vspace{-0.2in}
\end{table}

\subsection{Analyzing DTIM}

We now analyze the performance of DTIM across varying values of the diversity threshold ($\tau$) and word-stem lengths. 

\subsubsection{Diversity Threshold}

Table~\ref{tab:tau} analyzes the clustering quality trends of DTIM across varying values of $\tau$. The table suggests that DTIM is extremely robust to variations in diversity threshold, despite a slight preference towards values around $10$ and $20$. This suggests that a system designer looking to use DTIM need not carefully tune this parameter due to the inherent robustness. 

\subsubsection{Word Stem Length}

Given the nature of Malayalam language where the variations in word lengths are not as high as in English, it seemed very natural to use a word stem length of $2$. Moreover, very large words are uncommon in Malayalam. In our corpus, $50\%$ of words were found to contain five characters or less, the corresponding fraction being $71\%$ for upto seven characters. Our analysis of DTIM across variations in word-stem length, illustrated in Table~\ref{tab:stemlength} strongly supports this intuition with clustering quality peaking at stem-length of $2$ for $n \geq 2$. It is notable, however, that DTIM degrades gracefully on either side of $2$. Trends across different settings of word-stem length are interesting since they may provide clues about applicability for other languages with varying character granularities (e.g., each Chinese character corresponds to multiple characters in Latin-script). 

%\noindent{\bf DTIM Analysis: (i) Diversity Threshold:} We analyzed DTIM recall across varying values of the diversity threshold parameter, $\tau$. The recall was seen to be notably stable across widely varying values of $\tau$. In particular, values of $\tau$ in the range $[5,1000]$ were seen to yield final DTIM recall values of $>0.74$. \noindent{\bf (ii) Word Stem Length:} We observed that DTIM degrades gracefully with variations of the word-stem length from $2$. The recall for various word-stem lengths are as follows: $1:0.71$,  $2:0.76$, $3:0.72$, $4:0.61$. %Trends across different settings of word-stem length are interesting since they may provide clues about applicability for other languages with varying character granularities (e.g., each Chinese character corresponds to multiple characters in Latin-script). 
% Though we have used word-stems of length $2$ for the initialization in DTIM (Section~\ref{sec:divinit}), this can be varied. 
%The recall of the initialization (i.e., INIT) was seen to deteriorate from $0.69$ at $\tau=5$ to $0.65$ at $\tau=1000$; however, the DTIM iterations were evidently robust to such initialization quality deteriorations to a reasonable extent. 

\section{Discussion}

\subsection{Applicability to Other Languages} 

In contrast to earlier work focused on specific languages (e.g.,~\cite{koo2015unsupervised}) that use heuristics that are very specific to the language (such as expected patterns of consonants), DTIM heuristics are general-purpose in design. The only heuristic setting that is likely to require some tuning for applicability in other languages is the word-stem length. We expect the approach would generalize well to other Sanskrit-influenced Dravidian languages such as Kannada/Telugu. Unfortunately, we did not have any Kannada/Telugu/Kodava knowledge (Dravidian languages have largely disjoint speaker-populations) in our team, or access to labelled datasets in those languages (they are low-resource languages too); testing this on Kannada/Telugu/Tamil would be interesting future work. The method is expected to be less applicable to English, the language being significantly different and with potentially fewer transliterable words. 

\subsection{DTIM in an Application Context} 

Within any target application context, machine-labelled transliterable words (and their automatically generated transliterations) may need to manual screening for accountability reasons. The high accuracy at either ends of the ordering lends itself to be exploited in the following fashion; in lieu of employing experts to verify all labellings/transliterations, low-expertise volunteers (e.g., students) can be called in to verify labellings at the ends (top/bottom) of the lists with experts focusing on the middle (more ambiguous) part of the list; this frees up experts’ time as against a cross-spectrum expert-verification process, leading to direct cost savings. We also expect that DTIM followed by automatic transliterations of bottom-k words would aid in retrieval and machine translation scenarios.

%\vspace{-0.1in}

\section{Conclusions and Future Work}\label{sec:conclusions}

We considered the problem of unsupervised separation of transliterable and native words in Malayalam, a critical task in easing automated processing of Malayalam text in the company of other language text. We outlined a key observation on the differential diversity beyond word stems, and formulated an initialization heuristic that would coarsely separate native and transliterable words. We proposed the usage of probability distributions over character n-grams as a way of separately modelling native and transliterable words. We then formulated an iterative optimization method that alternatively refines the nativeness scorings and probability distributions. Our technique for the problem, DTIM, that encompasses the initialization and iterative refinement, was seen to significantly outperform other unsupervised baseline methods in our empirical study. This establishes DTIM as the preferred method for the task. We have also released our datasets and labeled subset to help aid future research on this and related tasks. 

%developed an iterative optimization method to arrive at an accurate nativeness ordering. Our empirical analysis establishes that our method, DTIM, outperforms unsupervised baseline methods by significant margins. This establishes DTIM as the state-of-the-art method for the task. %We have also released our datasets and labeled subset to help aid future research on this and related tasks. 

\subsection{Future Work}

The applicability of DTIM to other Dravidian languages is interesting to study. Due to our lack of familiarity with any other language in the family, we look forward to contacting other groups to further the generalizability study. While nativeness scoring improvements directly translate to reduction of effort for manual downstream processing, quantifying gains these bring about in translation and retrieval is interesting future work. Exploring the relationship/synergy of this task and Sandhi splitting~\cite{DBLP:conf/ijcnlp/NatarajanC11} would form another interesting direction for future work. Further, we would like to develop methods to separate out the two categories of transliterable words, viz., foreign language words and proper nouns. Such a method would enable a more fine-grained stratification of the vocabulary. 

Transliterable words are often within Malayalam used to refer to very topical content, for which suitable words are harder to find. Thus, transliterable words could be preferentially treated towards building rules in interpretable clustering~\cite{balachandran2012interpretable} and for modelling context in regex-oriented rule-based information extraction~\cite{murthy2012improving}. Transliterable words might also hold cues for detecting segment boundaries in conversational transcripts~\cite{kummamuru2009unsupervised,padmanabhan2007mining}. 

%and (ii) relationship/synergy of this task and Sandhi splitting~\cite{DBLP:conf/ijcnlp/NatarajanC11} form interesting future work directions. %Further, we would like to develop methods to separate out the two categories of transliterable words, viz., foreign language words and proper nouns. Such a method would enable a more fine-grained stratification of the vocabulary. 

%Another related problem that has not yet been addressed is splitting of 

%\begin{quote}
%\begin{verbatim}
%\bibliography{eacl2017}
%\bibliographystyle{eacl2017}
%\end{verbatim}
%\end{quote}

\bibliography{icon2017}
\bibliographystyle{acl}

\end{document}